\documentclass[draft]{article} 
\usepackage{lambek-fs,lscape,xypic,prooftree}

\title{Displacement Calculus}
\author{Glyn Morrill and Oriol Valent\'{\i}n}
\institution{Universitat Polit\`ecnica de Catalunya\\ and\\ Universitat Pompeu Fabra}

\newcommand{\techterm}[1]{\emph{#1}}
\newcommand{\disp}[1]{\enumsentence{#1}}
\newcommand{\product}{\mbox{$\bullet$}}
\newcommand{\bsl}{\mbox{$\backslash$}}
\newcommand{\inp}{\mbox{$^{\bullet}$}}
\newcommand{\out}{\mbox{$^{\circ}$}}
\newcommand{\scare}[1]{``#1''}
\newcommand{\mconj}{\mbox{$\,\otimes\,$}}
\newcommand{\mdisj}{\mbox{$\wp$}}

\newcommand{\tb}{\hspace*{0.25in}}
\newcommand{\yields}{\mbox{\ $\Rightarrow$\ }}
\newcommand{\AL}{\mbox{\bf L}}
\newcommand{\D}{\mbox{\bf D}}
\newcommand{\one}{\mbox{1}}
\newcommand{\add}{\mbox{$+$}}

\newcommand{\infix}{\mbox{$\downarrow$}}
\newcommand{\extract}{\mbox{$\uparrow$}}
\newcommand{\dprod}{\mbox{$\odot$}}
\newcommand{\sep}{\mbox{$[\,]$}}
\newcommand{\vect}[1]{\overrightarrow{#1}}
\newcommand{\lingform}[1]{\emph{#1}}
\newcommand{\vtab}{\ \vspace{2.5ex}\\}
\newcommand{\verbquote}[1]{`#1'}
\newcommand{\CN}{\mbox{\it CN}}

\newcommand{\mytheorem}[2]{\disp{{\bf Theorem} ({\em #1}).\\
\ \\#2}}
\newcommand{\mycorollary}[2]{\disp{{\bf Corollary} ({\em #1}).\\
\ \\#2}}

\newcommand{\prf}[1]{{\bf Proof}. #1\ $\Box$}

\spreaddiagramrows{-1.1pc}
\spreaddiagramcolumns{-1.1pc}
\newcommand{\fig}[3]{\begin{figure}\begin{center}#1\end{center}
\caption{#2}\label{#3}\end{figure}}
\newcommand{\rotfig}[3]{\begin{figure}\begin{center}\rotatebox{-90}{\tiny #1}\end{center}
\caption{#2}\label{#3}\end{figure}}
\newcommand{\netfig}[3]{\begin{figure}\begin{center}\spreaddiagramrows{-0.6pc}
\spreaddiagramcolumns{-0.6pc}
\small$\diagram#1\enddiagram$\end{center}
\caption{#2}\label{#3}\end{figure}}

\begin{document}

\maketitle

\begin{abstract}
The Lambek calculus \AL{} provides a foundation for categorial grammar
in the form of a logic of concatenation.
But natural language is characterized by dependencies which
may also be discontinuous.
In this paper we introduce the displacement calculus {\bf D},
a generalization of Lambek calculus,
which preserves the good proof-theoretic properties of the latter
while embracing discontinuiity and subsuming \AL.
We illustrate linguistic applications and prove
Cut-elimination,
the subformula property,
and decidability

\end{abstract}

\section{Introduction}

\cite{lambek:mathematics} applied mathematical logic to linguistics
in such a way that the analysis of a sentence is a proof.\footnote{The
research reported in the present paper
was supported by DGICYT project SESAAME-BAR
(TIN2008-06582-C03-01).}
This was the genesis of logical syntax,
a decade before the advent of logical semantics.
Once these applications of logic are born
they take on a life of their own,
for by comparison the rest seems \ldots{} illogical.
The Lambek calculus is a sequence logic without structural rules which enjoys
Cut-elimination, the subformula property, and decidability.
It is intuitionistic,
hence the standard Curry-Howard categorial
semantics. 
It is sound and complete with respect to interpretation by residuation in free semigroups.
But for all its elegance, as a logic of concatenation, the Lambek calculus can only analyse
displacement when the dependencies happen to be peripheral.
As a consequence it cannot account for the syntax and semantics of, for example:
\disp{
\begin{itemize}
\item
Discontinuous idioms (\lingform{Mary gave the man the cold shoulder}).
\item
Quantification (\lingform{John gave every book to Mary};
\lingform{Mary thinks someone left}; \lingform{Everyone loves someone}).
\item
VP ellipsis (\lingform{John slept before Mary did};
\lingform{John slept and Mary did too}).
\item
Medial extraction (\lingform{dog that Mary saw today}).
\item
Pied-piping (\lingform{mountain the painting of which by Cezanne John sold for
\$10,000,000}.)
\item
Appositive relativization (\lingform{John, who jogs, sneezed}).
\item
Parentheticals (\lingform{Fortunately, John has perseverance};
\lingform{John, fortunately, has perseverance};
\lingform{John has, fortunately, perseverance};
\lingform{John has perseverance, fortunately}).
\item
Gapping (\lingform{John studies logic, and Charles, phonetics}).
\item
Comparative subdeletion ({\em John ate more donuts than Mary bought bagels}).
\item
Reflexivization (\lingform{John sent himself flowers}).
\end{itemize}\label{dispphen}}
 
 In the decade of the 90s it seemed that a general methodology
 for obtaining more adequate categorial grammars might be
 to introduce families of residuated connectives for multiple modes
 of composition related by structural rules \citep{moortgat:handbook}: so-called
 multimodal categorial grammar. 
 But this paper marks a return to unimodal categorial grammar like the Lambek calculus,
 in that there is a single primitive mode of binary composition,
 namely concatenation;
 the modes of composition with respect to which the other connectives are specified
 are defined.
 Indeed,
 we present displacement
calculus which,
like the Lambek calculus,
is a sequence logic without structural rules which,
as we shall show here,
enjoys Cut-elimination,
the subformula property, and decidability.
Moreover, 
like the Lambek calculus it is intuitionistic,
and so supports the standard categorial Curry-Howard type-logical
semantics.
We shall show how it
provides basic analyses of all of the phenomena itemized in (\ref{dispphen}). 

In Section~\ref{dispsect} we define the calculus of displacement.
In Section~\ref{outsect} we give linguistic applications.
In Section~\ref{cutsect} we prove Cut-elimination,
and we conclude in Section~\ref{conclsect}.

\section{Displacement Calculus}

\label{dispsect}

 The types of the calculus of displacement \D{} classify strings over a vocabulary
 including a distinguished placeholder \one{} called the \techterm{separator}.
 The sort $i\in{\cal N}$ of a (discontinuous) string is the number of separators it contains and these
 punctuate it into $i+1$ maximal continuous substrings or \techterm{segments}.
 The types of \D{} are sorted into types ${\cal F}_i$ of sort $i$ by mutual recursion as follows:
 \disp{$
 \begin{array}[t]{rcllrcll}
 {\cal F}_j & := & {\cal F}_i\bsl{\cal F}_{i{+}j} & \mbox{under}\\
{\cal F}_i & := & {\cal F}_{i{+}j}/{\cal F}_j & \mbox{over}\\
 {\cal F}_{i{+}j} & := & {\cal F}_i\product{\cal F}_j & \mbox{product}\\
 {\cal F}_0 & := & I & \mbox{product unit}\\
{\cal F}_j & := & {\cal F}_{i{+}1}\infix_k{\cal F}_{i{+}j}, 1\le k\le i{+}1& \mbox{infix}\\
{\cal F}_{i{+}1} & := & {\cal F}_{i{+}j}\extract_k{\cal F}_j, 1\le k\le i{+}1 & \mbox{extract}\\
 {\cal F}_{i{+}j} & := & {\cal F}_{i{+}1}\dprod_k{\cal F}_j,1\le k\le i{+}1& \mbox{disc.\ product}\\
 {\cal F}_1 & := & J & \mbox{disc.\ prod.\ unit}\\
\end{array}$}
 Where $A$ is a type we call its sort $sA$.
 The set $\cal O$ of \techterm{configurations\/} is defined as follows,
 where $\Lambda$ is the empty string and \sep{} is the metalinguistic
 separator:
 \disp{$
 {\cal O} ::= \Lambda\ |\ \sep\ |\ {\cal F}_0\ |\ 
 {\cal F}_{i{+}1}\{\underbrace{{\cal O}: \ldots :{\cal O}}_{ i{+}1\ {\cal O}'s}\}\ |\  {\cal O}, {\cal O}$}
 Note that the configurations are of a new kind in which some type formulas, namely
 the type formulas of sort greater than one, label mother nodes rather than leaves, and have a number of
 immediate subconfigurations equal to their sort. 
 This signifies a discontinuous type intercalated by these subconfigurations.
 Thus $A\{\Delta_1: \ldots: \Delta_n\}$ interpreted syntactically is formed by strings
 $\alpha_0\add\beta_1\add\cdots\add\beta_n\add\alpha_n$
 where $\alpha_0\add\one\add\cdots\add\one\add\alpha_n \in A$ and
 $\beta_1\in \Delta_1, \ldots, \beta_n\in\Delta_n$.
 We call these types \techterm{hyperleaves\/} since in multimodal
 calculus they would be leaves.
 We call these new configurations \techterm{hyperconfigurations}.
 The sort of a (hyper)configuration is the number of separators it contains.
 A \techterm{hypersequent\/} $\Gamma\yields A$ comprises an  
 antecedent hyperconfiguration $\Gamma$
 of sort $i$ and a succedent type $A$ of sort $i$.
The \techterm{vector\/} \vect{A} of a type $A$ is defined by:
 \disp{$
 \vect{A} = \left\{
 \begin{array}{ll}
 A & \mbox{if\ } sA=0\\
 A\{\underbrace{\sep: \ldots: \sep}_{sA\ \sep's}\} & \mbox{if\ } sA>0
 \end{array}\right.$}
 Where $\Delta$ is a configuration of sort at least $k$ and $\Gamma$ 
 is a configuration, the \techterm{$k$-ary wrap}
 $\Delta|_k\Gamma$ signifies the configuration which is the result of
 replacing by $\Gamma$ the $k$th separator in $\Delta$.
 Where  $\Delta$ is a configuration of sort $i$ and $\Gamma_1, \ldots, \Gamma_i$ are configurations,
 the \techterm{generalized wrap\/} $\Delta\otimes\langle \Gamma_1, \ldots, \Gamma_i\rangle$ is the
 result of simultaneously replacing the successive separators in $\Delta$ by $\Gamma_1, \ldots, \Gamma_i$
 respectively.
In the hypersequent calculus we use a discontinuous distinguished hyperoccurrence notation 
$\Delta\langle\Gamma\rangle$
to refer to a configuration $\Delta$ and continuous subconfigurations  $\Delta_1, \ldots, \Delta_i$ and a discontinuous subconfiguration $\Gamma$ of sort $i$ such that
$\Gamma\otimes\langle \Delta_1, \ldots, \Delta_i\rangle$ is a continuous subconfiguration.
That is, where $\Gamma$ is of sort $i$,
$\Delta\langle\Gamma\rangle$ abbreviates
$\Delta(\Gamma\otimes\langle\Delta_1, \ldots, \Delta_i\rangle)$
where $\Delta(\ldots)$ is the usual distinguished occurrence notation.
Technically, whereas the usual distinguished occurrence
notation $\Delta(\Gamma)$ refers to a context
containing a \techterm{hole\/}
which is a leaf, in hypersequent calculus the distinguished hyperoccurrence
notation $\Delta\langle\Gamma\rangle$ refers to a context
containing a hole which may be a hyperleaf, a \techterm{hyperhole}.

The hypersequent calculus for the calculus of displacement is given
in Figure~\ref{hyper}. 
Observe that the rules for both the concatenating connectives
$\bsl, \product, /$ and the wrapping connectives $\infix_k, \dprod_k, \extract_k$
are just like the rules for Lambek calculus except for the vectorial notation
and hyperoccurrence notation;
the former are specified in relation to the primitive concatenation
represented by the sequent comma and the latter are specified in relation 
to the defined operations of $k$-ary wrap.

\begin{figure}
\begin{center}
$\prooftree
\justifies
\vect{A}\yields A
\using id
\endprooftree \tb
\prooftree
\Gamma\yields A \tb
\Delta\langle \vect{A}\rangle\yields B
\justifies
\Delta\langle\Gamma\rangle\yields B
\using Cut
\endprooftree$
\vtab
$\prooftree
\Gamma\yields A \tb
\Delta\langle\vect{C}\rangle\yields D
\justifies
\Delta\langle\Gamma, \vect{A\bsl C}\rangle\yields D
\using \bsl L
\endprooftree \tb
\prooftree
\vect{A}, \Gamma\yields C
\justifies
\Gamma\yields A\bsl C
\using \bsl R
\endprooftree$
\vtab
$\prooftree
\Gamma\yields B \tb
\Delta\langle\vect{C}\rangle\yields D
\justifies
\Delta\langle\vect{C/B}, \Gamma\rangle\yields D
\using / L
\endprooftree \tb
\prooftree
\Gamma, \vect{B}\yields C
\justifies
\Gamma\yields C/B
\using / R
\endprooftree$
\vtab
$\prooftree
\Delta\langle\vect{A}, \vect{B}\rangle\yields D
\justifies
\Delta\langle\vect{A\product B}\rangle\yields D
\using \product L
\endprooftree \tb
\prooftree
\Gamma_1\yields A\tb\Gamma_2\yields B
\justifies
\Gamma_1, \Gamma_2\yields A\product B
\using \product R
\endprooftree
$
\vtab
$\prooftree
\Delta\langle\Lambda\rangle\yields A
\justifies
\Delta\langle\vect{I}\rangle\yields A
\using IL
\endprooftree\tb
\prooftree
\justifies
\Lambda\yields I
\using IR
\endprooftree
$
\vtab
$\prooftree
\Gamma\yields A \tb
\Delta\langle\vect{C}\rangle\yields D
\justifies
\Delta\langle\Gamma|_k\vect{A\infix_k C}\rangle\yields D
\using \infix_k L
\endprooftree \tb
\prooftree
\vect{A}|_k\Gamma\yields C
\justifies
\Gamma\yields A\infix_k C
\using \infix_k R
\endprooftree$
\vtab
$\prooftree
\Gamma\yields B \tb
\Delta\langle\vect{C}\rangle\yields D
\justifies
\Delta\langle\vect{C\extract_k B}|_k\Gamma\rangle\yields D
\using \extract_k L
\endprooftree \tb
\prooftree
\Gamma|_k\vect{B}\yields C
\justifies
\Gamma\yields C\extract_k B
\using \extract_k R
\endprooftree$
\vtab
$\prooftree
\Delta\langle\vect{A}|_k\vect{B}\rangle\yields D
\justifies
\Delta\langle\vect{A\dprod_k B}\rangle\yields D
\using \dprod_k L
\endprooftree \tb
\prooftree
\Gamma_1\yields A\tb\Gamma_2\yields B
\justifies
\Gamma_1|_k\Gamma_2\yields A\dprod_k B
\using \dprod_k R
\endprooftree
$
\vtab
$\prooftree
\Delta\langle\sep\rangle\yields A
\justifies
\Delta\langle\vect{J}\rangle\yields A
\using JL
\endprooftree\tb
\prooftree
\justifies
\sep\yields J
\using JR
\endprooftree
$
\end{center}
\caption{Calculus of displacement \D}
\label{hyper}
\end{figure}

\section{Linguistic Applications}

\label{outsect}

A parser/theorem-prover for the displacement calculus has been
implemented in Prolog. 
In this section we give the analyses it produces for the
examples of (\ref{dispphen}). These are examples from Chapter~6
of \cite{morrill:oxford}.
There a very similar system called discontinuous Lambek calculus
is used with unary bridge and split operators and no nullary product units.
Here we use the displacement calculus which has the continuous
and discontinuous product units $I$ and $J$ instead of unary operators.
The lexicon for the analyses 
is as follows; we abbreviate $\infix_1, \dprod_1$ and $\extract_1$ as
$\infix, \dprod$ and $\extract$ respectively.

\disp{$
{\bf \$10,000,000}: N: {\it tenmilliondollars}\\
{\bf and}: (S\backslash S)/S: \lambda A\lambda B[{\it B}\wedge {\it A}]\\
{\bf and}:\\
\tb ((S{\uparrow}((N\backslash S)/N))\backslash (S{\uparrow}((N\backslash S)/N)))/((S{\uparrow}((N\backslash S)/N)){\odot}I):\\
\tb \lambda A\lambda B\lambda C[({\it B}\ {\it C})\wedge (\pi_1{\it A}\ {\it C})]\\
{\bf ate}: (N\backslash S)/N: {\it ate}\\
{\bf bagels}: {\it CN}: {\it bagels}\\
{\bf before}: ((N\backslash S)\backslash (N\backslash S))/S: \lambda A\lambda B\lambda C(({\it before}\ {\it A})\ ({\it B}\ {\it C}))\\
{\bf book}: {\it CN}: {\it book}\\
{\bf bought}: (N\backslash S)/N: {\it bought}\\
{\bf by}: ({\it CN}\backslash {\it CN})/N: {\it by}\\
{\bf cezanne}: N: {\it cezanne}\\
{\bf charles}: N: {\it c}\\
{\bf did}: (((N\backslash S){\uparrow}(N\backslash S))/(N\backslash S))\backslash ((N\backslash S){\uparrow}(N\backslash S)):\\
\tb \lambda A\lambda B(({\it A}\ {\it B})\ {\it B})\\
{\bf did}{+}{\bf too}: (((N\backslash S){\uparrow}(N\backslash S))/(N\backslash S))\backslash ((N\backslash S){\uparrow}(N\backslash S)):\\
\tb \lambda A\lambda B(({\it A}\ {\it B})\ {\it B})\\
{\bf dog}: {\it CN}: {\it dog}\\
{\bf donuts}: {\it CN}: {\it donuts}\\
{\bf every}: ((S{\uparrow}N){\downarrow}S)/{\it CN}: \lambda A\lambda B\forall C[({\it A}\ {\it C})\rightarrow ({\it B}\ {\it C})]\\
{\bf everyone}: (S{\uparrow}N){\downarrow}S: \lambda A\forall B[({\it person}\ {\it B})\rightarrow ({\it A}\ {\it B})]\\
{\bf flowers}: N: {\it flowers}\\
{\bf for}: {\it PP}/N: \lambda A{\it A}\\
{\bf fortunately}: (S{\uparrow}I){\downarrow}S: \lambda A({\it fortunately}\ ({\it A}\ {\it d}))\\
{\bf john}: N: {\it j}\\
{\bf gave}: (N\backslash S)/(N{\bullet}{\it PP}): \lambda A(({\it gave}\ \pi_2{\it A})\ \pi_1{\it A})\\
{\bf gave}{+}{\bf 1}{+}{\bf the}{+}{\bf cold}{+}{\bf shoulder}: (N\backslash S){\uparrow}N: {\it shunned}\\
{\bf has}: (N\backslash S)/N: {\it has}\\
{\bf himself}: ((N\backslash S){\uparrow}N){\downarrow}(N\backslash S): \lambda A\lambda B(({\it A}\ {\it B})\ {\it B})\\
{\bf jogs}: N\backslash S: {\it jogs}\\
{\bf left}: N\backslash S: {\it left}\\
{\bf logic}: N: {\it logic}\\
{\bf loves}: (N\backslash S)/N: {\it love}\\
{\bf man}: {\it CN}: {\it man}\\
{\bf mary}: N: {\it m}\\
{\bf more}:\\
\tb (S{\uparrow}(((S{\uparrow}N){\downarrow}S)/{\it CN})){\downarrow}(S/(({\it CP}{\uparrow}(((S{\uparrow}N){\downarrow}S)/{\it CN})){\odot}I)):
\tb \lambda A\lambda B[|\lambda C({\it A}\ \lambda D\lambda E[({\it D}\ {\it C})\wedge ({\it E}\ {\it C})])|>|\lambda C(\pi_1{\it B}\ \lambda D\lambda E\\
\tb\tb [({\it D}\ {\it C})\wedge ({\it E}\ {\it C})])|]\\
{\bf mountain}: {\it CN}: {\it mountain}\\
{\bf painting}: {\it CN}: {\it painting}\\
{\bf perseverance}: N: {\it perseverance}\\
{\bf phonetics}: N: {\it phonetics}\\
{\bf of}: ({\it CN}\backslash {\it CN})/N: {\it of}\\
{\bf slept}: N\backslash S: {\it slept}\\
{\bf saw}: (N\backslash S)/N: {\it saw}\\
{\bf sent}: (N\backslash S)/(N{\bullet}N): \lambda A(({\it sent}\ \pi_1{\it A})\ \pi_2{\it A})\\
{\bf sneezed}: N\backslash S: {\it sneezed}\\
{\bf sold}: (N\backslash S)/(N{\bullet}{\it PP}): \lambda A(({\it sold}\ \pi_2{\it A})\ \pi_1{\it A})\\
{\bf someone}: (S{\uparrow}N){\downarrow}S: \lambda A\exists B[({\it person}\ {\it B})\wedge ({\it A}\ {\it B})]\\
{\bf studies}: (N\backslash S)/N: {\it studies}\\
{\bf than}: {\it CP}/S: \lambda A{\it A}\\
{\bf that}: ({\it CN}\backslash {\it CN})/((S{\uparrow}N){\odot}I): \lambda A\lambda B\lambda C[({\it B}\ {\it C})\wedge (\pi_1{\it A}\ {\it C})]\\
{\bf the}: N/{\it CN}: \iota \\
{\bf thinks}: (N\backslash S)/S: {\it thinks}\\
{\bf to}: {\it PP}/N: \lambda A{\it A}\\
{\bf today}: (N\backslash S)\backslash (N\backslash S): \lambda A\lambda B({\it today}\ ({\it A}\ {\it B}))\\
{\bf which}: (N{\uparrow}N){\downarrow}(({\it CN}\backslash {\it CN})/((S{\uparrow}N){\odot}I)):\\
\tb \lambda A\lambda B\lambda C\lambda D[({\it C}\ {\it D})\wedge (\pi_1{\it B}\ ({\it A}\ {\it D}))]\\
{\bf who}: (N\backslash ((S{\uparrow}N){\downarrow}S))/((S{\uparrow}N){\odot}I)\\
\tb \lambda A\lambda B\lambda C[(\pi_1{\it A}\ {\it B})\wedge ({\it C}\ {\it B})]\}$}

The phenomena itemized in (\ref{dispphen}) are considered in the following subsections.

\subsection{Discontinuous Idioms}

Our first example is of a discontinuous idiom, where the lexicon has to
assign \lingform{give \ldots the cold shoulder} a non-composicional
meaning \verbquote{shun}:
\disp{
${\bf mary}{+}{\bf gave}{+}{\bf the}{+}{\bf man}{+}{\bf the}{+}{\bf cold}{+}{\bf shoulder}: S$}
Lexical insertion yields the following sequent,
which is labelled with the lexical semantics:
\disp{
$N: {\it m}, (N\backslash S){\uparrow}N\{N/{\it CN}: \iota , {\it CN}: {\it man}\}: {\it shunned}\ \Rightarrow\ S$}
This has a proof as follows.
\disp{
\prooftree
\prooftree
\prooftree
\justifies
{\it CN}\ \Rightarrow\ {\it CN}
\endprooftree
\prooftree
\justifies
N\ \Rightarrow\ N
\endprooftree
\justifies
N/{\it CN}, {\it CN}\ \Rightarrow\ N
\using /L
\endprooftree
\prooftree
\prooftree
\justifies
N\ \Rightarrow\ N
\endprooftree
\prooftree
\justifies
S\ \Rightarrow\ S
\endprooftree
\justifies
N, N\backslash S\ \Rightarrow\ S
\using \backslash L
\endprooftree
\justifies
N, (N\backslash S){\uparrow}N\{N/{\it CN}, {\it CN}\}\ \Rightarrow\ S
\using {\uparrow}L
\endprooftree}
This delivers the semantics:
\disp{
$(({\it shunned}\ (\iota \ {\it man}))\ {\it m})$}

\subsection{Quantification}

Lambek categorial grammar can analyse a subject quantifier phrase
by assigning it type $S/(N\bsl S)$. To obtain an object quantifier phrase
it requires another type $(S/N)\bsl S)$. But to analyse an example as follows
with a medial quantifier phrase would require still another type.
\disp{
${\bf john}{+}{\bf gave}{+}{\bf every}{+}{\bf book}{+}{\bf to}{+}{\bf mary}: S$}
Our treatment on the other hand requires just a single type $(S\extract N)\infix S$
for all quantifier phrase positions. Lexical insertion for this example
yields the following semantically labelled sequent:
\disp{
$N: {\it j}, (N\backslash S)/(N{\bullet}{\it PP}): \lambda A(({\it gave}\ \pi_2{\it A})\ \pi_1{\it A}), ((S{\uparrow}N){\downarrow}S)/{\it CN}: \lambda A\lambda B\forall C[({\it A}\ {\it C})\rightarrow ({\it B}\ {\it C})], {\it CN}: {\it book}, {\it PP}/N: \lambda A{\it A}, N: {\it m}\ \Rightarrow\ S$}
This is proved as follows:
\disp{\scriptsize
\prooftree
\prooftree
\justifies
{\it CN}\ \Rightarrow\ {\it CN}
\endprooftree
\prooftree
\prooftree
\prooftree
\prooftree
\prooftree
\justifies
N\ \Rightarrow\ N
\endprooftree
\prooftree
\prooftree
\justifies
N\ \Rightarrow\ N
\endprooftree
\prooftree
\justifies
{\it PP}\ \Rightarrow\ {\it PP}
\endprooftree
\justifies
{\it PP}/N, N\ \Rightarrow\ {\it PP}
\using /L
\endprooftree
\justifies
N, {\it PP}/N, N\ \Rightarrow\ N{\bullet}{\it PP}
\using \bullet R
\endprooftree
\prooftree
\prooftree
\justifies
N\ \Rightarrow\ N
\endprooftree
\prooftree
\justifies
S\ \Rightarrow\ S
\endprooftree
\justifies
N, N\backslash S\ \Rightarrow\ S
\using \backslash L
\endprooftree
\justifies
N, (N\backslash S)/(N{\bullet}{\it PP}), N, {\it PP}/N, N\ \Rightarrow\ S
\using /L
\endprooftree
\justifies
N, (N\backslash S)/(N{\bullet}{\it PP}), [\;], {\it PP}/N, N\ \Rightarrow\ S{\uparrow}N
\using {\uparrow}R
\endprooftree
\prooftree
\justifies
S\ \Rightarrow\ S
\endprooftree
\justifies
N, (N\backslash S)/(N{\bullet}{\it PP}), (S{\uparrow}N){\downarrow}S, {\it PP}/N, N\ \Rightarrow\ S
\using {\downarrow}L
\endprooftree
\justifies
N, (N\backslash S)/(N{\bullet}{\it PP}), ((S{\uparrow}N){\downarrow}S)/{\it CN}, {\it CN}, {\it PP}/N, N\ \Rightarrow\ S
\using /L
\endprooftree}
The semantics is thus:
\disp{
$\forall C[({\it book}\ {\it C})\rightarrow ((({\it gave}\ {\it m})\ {\it C})\ {\it j})]$}

The next example exhibits de re/de dicto ambiguity:
\disp{
${\bf mary}{+}{\bf thinks}{+}{\bf someone}{+}{\bf left}: S$}
Mary's thoughts could be specifically directed towards a particular
person, or concern a non-specific person. 
Lexical lookup yields the following:
\disp{
$N: {\it m}, (N\backslash S)/S: {\it thinks}, (S{\uparrow}N){\downarrow}S: \lambda A\exists B[({\it person}\ {\it B})\wedge ({\it A}\ {\it B})], N\backslash S: {\it left}\ \Rightarrow\ S$}
The non-specific
derivation and semantics are thus:
\disp{\small
\prooftree
\prooftree
\prooftree
\prooftree
\prooftree
\justifies
N\ \Rightarrow\ N
\endprooftree
\prooftree
\justifies
S\ \Rightarrow\ S
\endprooftree
\justifies
N, N\backslash S\ \Rightarrow\ S
\using \backslash L
\endprooftree
\justifies
[\;], N\backslash S\ \Rightarrow\ S{\uparrow}N
\using {\uparrow}R
\endprooftree
\prooftree
\justifies
S\ \Rightarrow\ S
\endprooftree
\justifies
(S{\uparrow}N){\downarrow}S, N\backslash S\ \Rightarrow\ S
\using {\downarrow}L
\endprooftree
\prooftree
\prooftree
\justifies
N\ \Rightarrow\ N
\endprooftree
\prooftree
\justifies
S\ \Rightarrow\ S
\endprooftree
\justifies
N, N\backslash S\ \Rightarrow\ S
\using \backslash L
\endprooftree
\justifies
N, (N\backslash S)/S, (S{\uparrow}N){\downarrow}S, N\backslash S\ \Rightarrow\ S
\using /L
\endprooftree}
\disp{
$(({\it thinks}\ \exists B[({\it person}\ {\it B})\wedge ({\it left}\ {\it B})])\ {\it m})$}
The specific derivation and semantics are:
\disp{\small
\prooftree
\prooftree
\prooftree
\prooftree
\prooftree
\justifies
N\ \Rightarrow\ N
\endprooftree
\prooftree
\justifies
S\ \Rightarrow\ S
\endprooftree
\justifies
N, N\backslash S\ \Rightarrow\ S
\using \backslash L
\endprooftree
\prooftree
\prooftree
\justifies
N\ \Rightarrow\ N
\endprooftree
\prooftree
\justifies
S\ \Rightarrow\ S
\endprooftree
\justifies
N, N\backslash S\ \Rightarrow\ S
\using \backslash L
\endprooftree
\justifies
N, (N\backslash S)/S, N, N\backslash S\ \Rightarrow\ S
\using /L
\endprooftree
\justifies
N, (N\backslash S)/S, [\;], N\backslash S\ \Rightarrow\ S{\uparrow}N
\using {\uparrow}R
\endprooftree
\prooftree
\justifies
S\ \Rightarrow\ S
\endprooftree
\justifies
N, (N\backslash S)/S, (S{\uparrow}N){\downarrow}S, N\backslash S\ \Rightarrow\ S
\using {\downarrow}L
\endprooftree}
\disp{
$\exists B[({\it person}\ {\it B})\wedge (({\it thinks}\ ({\it left}\ {\it B}))\ {\it m})]$}

Consider the classic example of quantifier scope ambiguity:
\disp{
${\bf everyone}{+}{\bf loves}{+}{\bf someone}: S$}
Lexical lookup yields:
\disp{
$(S{\uparrow}N){\downarrow}S: \lambda A\forall B[({\it person}\ {\it B})\rightarrow ({\it A}\ {\it B})], (N\backslash S)/N: {\it love},\\ (S{\uparrow}N){\downarrow}S: \lambda A\exists B[({\it person}\ {\it B})\wedge ({\it A}\ {\it B})]\ \Rightarrow\ S$}
In the object wide scope analysis the object quantifier phrase is processed
first top-down:
\disp{\small
\prooftree
\prooftree
\prooftree
\prooftree
\prooftree
\prooftree
\justifies
N\ \Rightarrow\ N
\endprooftree
\prooftree
\prooftree
\justifies
N\ \Rightarrow\ N
\endprooftree
\prooftree
\justifies
S\ \Rightarrow\ S
\endprooftree
\justifies
N, N\backslash S\ \Rightarrow\ S
\using \backslash L
\endprooftree
\justifies
N, (N\backslash S)/N, N\ \Rightarrow\ S
\using /L
\endprooftree
\justifies
[\;], (N\backslash S)/N, N\ \Rightarrow\ S{\uparrow}N
\using {\uparrow}R
\endprooftree
\prooftree
\justifies
S\ \Rightarrow\ S
\endprooftree
\justifies
(S{\uparrow}N){\downarrow}S, (N\backslash S)/N, N\ \Rightarrow\ S
\using {\downarrow}L
\endprooftree
\justifies
(S{\uparrow}N){\downarrow}S, (N\backslash S)/N, [\;]\ \Rightarrow\ S{\uparrow}N
\using {\uparrow}R
\endprooftree
\prooftree
\justifies
S\ \Rightarrow\ S
\endprooftree
\justifies
(S{\uparrow}N){\downarrow}S, (N\backslash S)/N, (S{\uparrow}N){\downarrow}S\ \Rightarrow\ S
\using {\downarrow}L
\endprooftree}
\disp{
$\exists B[({\it person}\ {\it B})\wedge \forall E[({\it person}\ {\it E})\rightarrow (({\it love}\ {\it B})\ {\it E})]]$}
In the subject wide scope analysis the subject quantifier phrase is processed
first top-down:
\disp{\small
\prooftree
\prooftree
\prooftree
\prooftree
\prooftree
\prooftree
\justifies
N\ \Rightarrow\ N
\endprooftree
\prooftree
\prooftree
\justifies
N\ \Rightarrow\ N
\endprooftree
\prooftree
\justifies
S\ \Rightarrow\ S
\endprooftree
\justifies
N, N\backslash S\ \Rightarrow\ S
\using \backslash L
\endprooftree
\justifies
N, (N\backslash S)/N, N\ \Rightarrow\ S
\using /L
\endprooftree
\justifies
N, (N\backslash S)/N, [\;]\ \Rightarrow\ S{\uparrow}N
\using {\uparrow}R
\endprooftree
\prooftree
\justifies
S\ \Rightarrow\ S
\endprooftree
\justifies
N, (N\backslash S)/N, (S{\uparrow}N){\downarrow}S\ \Rightarrow\ S
\using {\downarrow}L
\endprooftree
\justifies
[\;], (N\backslash S)/N, (S{\uparrow}N){\downarrow}S\ \Rightarrow\ S{\uparrow}N
\using {\uparrow}R
\endprooftree
\prooftree
\justifies
S\ \Rightarrow\ S
\endprooftree
\justifies
(S{\uparrow}N){\downarrow}S, (N\backslash S)/N, (S{\uparrow}N){\downarrow}S\ \Rightarrow\ S
\using {\downarrow}L
\endprooftree}
\disp{
$\forall B[({\it person}\ {\it B})\rightarrow \exists E[({\it person}\ {\it E})\wedge (({\it love}\ {\it E})\ {\it B})]]$}

\subsection{VP Ellipsis}

In VP ellipsis an auxiliary such as \lingform{did\/} receives its interpretion from an
antecedent verb phrase:
\disp{
${\bf john}{+}{\bf slept}{+}{\bf before}{+}{\bf mary}{+}{\bf did}: S$}
Lexical lookup for this example yields the following labelled sequent.
\disp{
$N: {\it j}, N\backslash S: {\it slept},\\ ((N\backslash S)\backslash (N\backslash S))/S: \lambda A\lambda B\lambda C(({\it before}\ {\it A})\ ({\it B}\ {\it C})), N: {\it m},\\ (((N\backslash S){\uparrow}(N\backslash S))/(N\backslash S))\backslash ((N\backslash S){\uparrow}(N\backslash S)): \lambda A\lambda B(({\it A}\ {\it B})\ {\it B})\\ \Rightarrow\ S$}
This has the proof given in Figure~\ref{jsbmd}.
\begin{figure}\tiny
$$
\prooftree
\prooftree
\prooftree
\prooftree
\prooftree
\prooftree
\prooftree
\justifies
N\ \Rightarrow\ N
\endprooftree
\prooftree
\justifies
S\ \Rightarrow\ S
\endprooftree
\justifies
N, N\backslash S\ \Rightarrow\ S
\using \backslash L
\endprooftree
\prooftree
\prooftree
\prooftree
\prooftree
\justifies
N\ \Rightarrow\ N
\endprooftree
\prooftree
\justifies
S\ \Rightarrow\ S
\endprooftree
\justifies
N, N\backslash S\ \Rightarrow\ S
\using \backslash L
\endprooftree
\justifies
N\backslash S\ \Rightarrow\ N\backslash S
\using \backslash R
\endprooftree
\prooftree
\prooftree
\justifies
N\ \Rightarrow\ N
\endprooftree
\prooftree
\justifies
S\ \Rightarrow\ S
\endprooftree
\justifies
N, N\backslash S\ \Rightarrow\ S
\using \backslash L
\endprooftree
\justifies
N, N\backslash S, (N\backslash S)\backslash (N\backslash S)\ \Rightarrow\ S
\using \backslash L
\endprooftree
\justifies
N, N\backslash S, ((N\backslash S)\backslash (N\backslash S))/S, N, N\backslash S\ \Rightarrow\ S
\using /L
\endprooftree
\justifies
N\backslash S, ((N\backslash S)\backslash (N\backslash S))/S, N, N\backslash S\ \Rightarrow\ N\backslash S
\using \backslash R
\endprooftree
\justifies
[\;], ((N\backslash S)\backslash (N\backslash S))/S, N, N\backslash S\ \Rightarrow\ (N\backslash S){\uparrow}(N\backslash S)
\using {\uparrow}R
\endprooftree
\justifies
[\;], ((N\backslash S)\backslash (N\backslash S))/S, N\ \Rightarrow\ ((N\backslash S){\uparrow}(N\backslash S))/(N\backslash S)
\using /R
\endprooftree
\prooftree
\prooftree
\prooftree
\prooftree
\justifies
N\ \Rightarrow\ N
\endprooftree
\prooftree
\justifies
S\ \Rightarrow\ S
\endprooftree
\justifies
N, N\backslash S\ \Rightarrow\ S
\using \backslash L
\endprooftree
\justifies
N\backslash S\ \Rightarrow\ N\backslash S
\using \backslash R
\endprooftree
\prooftree
\prooftree
\justifies
N\ \Rightarrow\ N
\endprooftree
\prooftree
\justifies
S\ \Rightarrow\ S
\endprooftree
\justifies
N, N\backslash S\ \Rightarrow\ S
\using \backslash L
\endprooftree
\justifies
N, (N\backslash S){\uparrow}(N\backslash S)\{N\backslash S\}\ \Rightarrow\ S
\using {\uparrow}L
\endprooftree
\justifies
N, N\backslash S, ((N\backslash S)\backslash (N\backslash S))/S, N, (((N\backslash S){\uparrow}(N\backslash S))/(N\backslash S))\backslash ((N\backslash S){\uparrow}(N\backslash S))\ \Rightarrow\ S
\using \backslash L
\endprooftree
$$
\caption{\lingform{John slept before Mary did}}
\label{jsbmd}
\end{figure}
The semantics is:
\disp{
$(({\it before}\ ({\it slept}\ {\it m}))\ ({\it slept}\ {\it j}))$}

By way of a second example consider:
\disp{
${\bf john}{+}{\bf slept}{+}{\bf and}{+}{\bf mary}{+}{\bf did}{+}{\bf too}: S$}
Lexical lookup yields:
\disp{
$N: {\it j}, N\backslash S: {\it slept}, (S\backslash S)/S: \lambda A\lambda B[{\it B}\wedge {\it A}], N: {\it m},\\ (((N\backslash S){\uparrow}(N\backslash S))/(N\backslash S))\backslash ((N\backslash S){\uparrow}(N\backslash S)): \lambda A\lambda B(({\it A}\ {\it B})\ {\it B})\\ \Rightarrow\ S$}
This has the proof given in Figure~\ref{jsamdt}.
\begin{figure}\tiny
$$
\prooftree
\prooftree
\prooftree
\prooftree
\prooftree
\prooftree
\justifies
N\ \Rightarrow\ N
\endprooftree
\prooftree
\prooftree
\prooftree
\justifies
N\ \Rightarrow\ N
\endprooftree
\prooftree
\justifies
S\ \Rightarrow\ S
\endprooftree
\justifies
N, N\backslash S\ \Rightarrow\ S
\using \backslash L
\endprooftree
\prooftree
\prooftree
\justifies
S\ \Rightarrow\ S
\endprooftree
\prooftree
\justifies
S\ \Rightarrow\ S
\endprooftree
\justifies
S, S\backslash S\ \Rightarrow\ S
\using \backslash L
\endprooftree
\justifies
S, (S\backslash S)/S, N, N\backslash S\ \Rightarrow\ S
\using /L
\endprooftree
\justifies
N, N\backslash S, (S\backslash S)/S, N, N\backslash S\ \Rightarrow\ S
\using \backslash L
\endprooftree
\justifies
N\backslash S, (S\backslash S)/S, N, N\backslash S\ \Rightarrow\ N\backslash S
\using \backslash R
\endprooftree
\justifies
[\;], (S\backslash S)/S, N, N\backslash S\ \Rightarrow\ (N\backslash S){\uparrow}(N\backslash S)
\using {\uparrow}R
\endprooftree
\justifies
[\;], (S\backslash S)/S, N\ \Rightarrow\ ((N\backslash S){\uparrow}(N\backslash S))/(N\backslash S)
\using /R
\endprooftree
\prooftree
\prooftree
\prooftree
\prooftree
\justifies
N\ \Rightarrow\ N
\endprooftree
\prooftree
\justifies
S\ \Rightarrow\ S
\endprooftree
\justifies
N, N\backslash S\ \Rightarrow\ S
\using \backslash L
\endprooftree
\justifies
N\backslash S\ \Rightarrow\ N\backslash S
\using \backslash R
\endprooftree
\prooftree
\prooftree
\justifies
N\ \Rightarrow\ N
\endprooftree
\prooftree
\justifies
S\ \Rightarrow\ S
\endprooftree
\justifies
N, N\backslash S\ \Rightarrow\ S
\using \backslash L
\endprooftree
\justifies
N, (N\backslash S){\uparrow}(N\backslash S)\{N\backslash S\}\ \Rightarrow\ S
\using {\uparrow}L
\endprooftree
\justifies
N, N\backslash S, (S\backslash S)/S, N, (((N\backslash S){\uparrow}(N\backslash S))/(N\backslash S))\backslash ((N\backslash S){\uparrow}(N\backslash S))\ \Rightarrow\ S
\using \backslash L
\endprooftree
$$
\caption{\lingform{John slept and Mary did too}}
\label{jsamdt}
\end{figure}
The semantics is:
\disp{
$[({\it slept}\ {\it j})\wedge ({\it slept}\ {\it m})]$}

\subsection{Medial Extraction}

Lambek categorial grammar can characterize subject relativization
with a relative pronoun type $(\CN\bsl\CN)/(N\bsl S)$ and clause-final
object relativization with a relative pronoun type $(\CN\bsl\CN)/(S/N)$,
but neither of these suffice for medial relativization such as the following:
\disp{
${\bf dog}{+}{\bf that}{+}{\bf mary}{+}{\bf saw}{+}{\bf today}: {\it CN}$}
Extraction from all positions is obtained with our displacement calculus
type, for which lexical lookup yields:
\disp{
${\it CN}: {\it dog}, ({\it CN}\backslash {\it CN})/((S{\uparrow}N){\odot}I): \lambda A\lambda B\lambda C[({\it B}\ {\it C})\wedge (\pi_1{\it A}\ {\it C})],\\ N: {\it m}, (N\backslash S)/N: {\it saw}, (N\backslash S)\backslash (N\backslash S): \lambda A\lambda B({\it today}\ ({\it A}\ {\it B}))\ \Rightarrow\ {\it CN}$}
The proof analysis is:
\disp{\tiny
\prooftree
\prooftree
\prooftree
\prooftree
\prooftree
\justifies
N\ \Rightarrow\ N
\endprooftree
\prooftree
\prooftree
\prooftree
\prooftree
\justifies
N\ \Rightarrow\ N
\endprooftree
\prooftree
\justifies
S\ \Rightarrow\ S
\endprooftree
\justifies
N, N\backslash S\ \Rightarrow\ S
\using \backslash L
\endprooftree
\justifies
N\backslash S\ \Rightarrow\ N\backslash S
\using \backslash R
\endprooftree
\prooftree
\prooftree
\justifies
N\ \Rightarrow\ N
\endprooftree
\prooftree
\justifies
S\ \Rightarrow\ S
\endprooftree
\justifies
N, N\backslash S\ \Rightarrow\ S
\using \backslash L
\endprooftree
\justifies
N, N\backslash S, (N\backslash S)\backslash (N\backslash S)\ \Rightarrow\ S
\using \backslash L
\endprooftree
\justifies
N, (N\backslash S)/N, N, (N\backslash S)\backslash (N\backslash S)\ \Rightarrow\ S
\using /L
\endprooftree
\justifies
N, (N\backslash S)/N, [\;], (N\backslash S)\backslash (N\backslash S)\ \Rightarrow\ S{\uparrow}N
\using {\uparrow}R
\endprooftree
\prooftree
\justifies
\ \Rightarrow\ I
\using IR
\endprooftree
\justifies
N, (N\backslash S)/N, (N\backslash S)\backslash (N\backslash S)\ \Rightarrow\ (S{\uparrow}N){\odot}I
\using {\odot}R
\endprooftree
\prooftree
\prooftree
\justifies
{\it CN}\ \Rightarrow\ {\it CN}
\endprooftree
\prooftree
\justifies
{\it CN}\ \Rightarrow\ {\it CN}
\endprooftree
\justifies
{\it CN}, {\it CN}\backslash {\it CN}\ \Rightarrow\ {\it CN}
\using \backslash L
\endprooftree
\justifies
{\it CN}, ({\it CN}\backslash {\it CN})/((S{\uparrow}N){\odot}I), N, (N\backslash S)/N, (N\backslash S)\backslash (N\backslash S)\ \Rightarrow\ {\it CN}
\using /L
\endprooftree}
This delivers semantics:
\disp{
$\lambda C[({\it dog}\ {\it C})\wedge ({\it today}\ (({\it saw}\ {\it C})\ {\it m}))]$}

\subsection{Pied-Piping}

In pied-piping a relative pronoun is accompanied by further material
from the extraction site:
\disp{
${\bf mountain}{+}{\bf the}{+}{\bf painting}{+}{\bf of}{+}{\bf which}{+}{\bf by}{+}{\bf cezanne}{+}{\bf john}{+}\\{\bf sold}{+}{\bf for}{+}{\bf \$10,000,000}: {\it CN}$}
Thje type we use for this example subsumes that of the previous subsection
since the latter is derivable from the former, so the lexicon only requires
the type employed in this lexical
lookup:
\disp{
${\it CN}: {\it mountain}, N/{\it CN}: \iota , {\it CN}: {\it painting}, ({\it CN}\backslash {\it CN})/N: {\it of}, (N{\uparrow}N){\downarrow}(({\it CN}\backslash {\it CN})/((S{\uparrow}N){\odot}I)): \lambda A\lambda B\lambda C\lambda D[({\it C}\ {\it D})\wedge\\ (\pi_1{\it B}\ ({\it A}\ {\it D}))], ({\it CN}\backslash {\it CN})/N: {\it by}, N: {\it cezanne}, N: {\it j},\\ (N\backslash S)/(N{\bullet}{\it PP}): \lambda A(({\it sold}\ \pi_2{\it A})\ \pi_1{\it A}), {\it PP}/N: \lambda A{\it A},\\ N: {\it tenmilliondollars}\ \Rightarrow\ {\it CN}$}
The derivation is given in Figure~\ref{cezanne}.
\begin{figure}
$$
\rotatebox{-90}{\tiny
\prooftree
\prooftree
\prooftree
\prooftree
\prooftree
\justifies
N\ \Rightarrow\ N
\endprooftree
\prooftree
\prooftree
\justifies
{\it CN}\ \Rightarrow\ {\it CN}
\endprooftree
\prooftree
\prooftree
\justifies
N\ \Rightarrow\ N
\endprooftree
\prooftree
\prooftree
\justifies
{\it CN}\ \Rightarrow\ {\it CN}
\endprooftree
\prooftree
\justifies
{\it CN}\ \Rightarrow\ {\it CN}
\endprooftree
\justifies
{\it CN}, {\it CN}\backslash {\it CN}\ \Rightarrow\ {\it CN}
\using \backslash L
\endprooftree
\justifies
{\it CN}, ({\it CN}\backslash {\it CN})/N, N\ \Rightarrow\ {\it CN}
\using /L
\endprooftree
\justifies
{\it CN}, {\it CN}\backslash {\it CN}, ({\it CN}\backslash {\it CN})/N, N\ \Rightarrow\ {\it CN}
\using \backslash L
\endprooftree
\justifies
{\it CN}, ({\it CN}\backslash {\it CN})/N, N, ({\it CN}\backslash {\it CN})/N, N\ \Rightarrow\ {\it CN}
\using /L
\endprooftree
\prooftree
\justifies
N\ \Rightarrow\ N
\endprooftree
\justifies
N/{\it CN}, {\it CN}, ({\it CN}\backslash {\it CN})/N, N, ({\it CN}\backslash {\it CN})/N, N\ \Rightarrow\ N
\using /L
\endprooftree
\justifies
N/{\it CN}, {\it CN}, ({\it CN}\backslash {\it CN})/N, [\;], ({\it CN}\backslash {\it CN})/N, N\ \Rightarrow\ N{\uparrow}N
\using {\uparrow}R
\endprooftree
\prooftree
\prooftree
\prooftree
\prooftree
\prooftree
\prooftree
\justifies
N\ \Rightarrow\ N
\endprooftree
\prooftree
\prooftree
\justifies
N\ \Rightarrow\ N
\endprooftree
\prooftree
\justifies
{\it PP}\ \Rightarrow\ {\it PP}
\endprooftree
\justifies
{\it PP}/N, N\ \Rightarrow\ {\it PP}
\using /L
\endprooftree
\justifies
N, {\it PP}/N, N\ \Rightarrow\ N{\bullet}{\it PP}
\using \bullet R
\endprooftree
\prooftree
\prooftree
\justifies
N\ \Rightarrow\ N
\endprooftree
\prooftree
\justifies
S\ \Rightarrow\ S
\endprooftree
\justifies
N, N\backslash S\ \Rightarrow\ S
\using \backslash L
\endprooftree
\justifies
N, (N\backslash S)/(N{\bullet}{\it PP}), N, {\it PP}/N, N\ \Rightarrow\ S
\using /L
\endprooftree
\justifies
N, (N\backslash S)/(N{\bullet}{\it PP}), [\;], {\it PP}/N, N\ \Rightarrow\ S{\uparrow}N
\using {\uparrow}R
\endprooftree
\prooftree
\justifies
\ \Rightarrow\ I
\using IR
\endprooftree
\justifies
N, (N\backslash S)/(N{\bullet}{\it PP}), {\it PP}/N, N\ \Rightarrow\ (S{\uparrow}N){\odot}I
\using {\odot}R
\endprooftree
\prooftree
\prooftree
\justifies
{\it CN}\ \Rightarrow\ {\it CN}
\endprooftree
\prooftree
\justifies
{\it CN}\ \Rightarrow\ {\it CN}
\endprooftree
\justifies
{\it CN}, {\it CN}\backslash {\it CN}\ \Rightarrow\ {\it CN}
\using \backslash L
\endprooftree
\justifies
{\it CN}, ({\it CN}\backslash {\it CN})/((S{\uparrow}N){\odot}I), N, (N\backslash S)/(N{\bullet}{\it PP}), {\it PP}/N, N\ \Rightarrow\ {\it CN}
\using /L
\endprooftree
\justifies
{\it CN}, N/{\it CN}, {\it CN}, ({\it CN}\backslash {\it CN})/N, (N{\uparrow}N){\downarrow}(({\it CN}\backslash {\it CN})/((S{\uparrow}N){\odot}I)), ({\it CN}\backslash {\it CN})/N, N, N, (N\backslash S)/(N{\bullet}{\it PP}), {\it PP}/N, N\ \Rightarrow\ {\it CN}
\using {\downarrow}L
\endprooftree}
$$
\caption{\lingform{mountain the painting of which by Cezanne John sold for \$10,000,000}}
\label{cezanne}
\end{figure}
This assigns semantics:
\disp{
$\lambda D[({\it mountain}\ {\it D})\wedge ((({\it sold}\ {\it tenmilliondollars})\ (\iota \ (({\it by}\ {\it cezanne})\\\ (({\it of}\ {\it D})\ {\it painting}))))\ {\it j})]$}

\subsection{Appositive Relativization}

In appositive relativization the head modified by the relative clause is
a noun phrase and the predication of the body of the relative clause to this head
is conjoined to the propositional content of the head interpreted in the embedding
sentence. Our example is:
\disp{${\bf john}{+}{\bf who}{+}{\bf jogs}{+}{\bf sneezed}: S$}
Lexical lookup yields:
\disp{
$N: {\it j}, (N\backslash ((S{\uparrow}N){\downarrow}S))/((S{\uparrow}N){\odot}I): \lambda A\lambda B\lambda C[(\pi_1{\it A}\ {\it B})\wedge\\ ({\it C}\ {\it B})], N\backslash S: {\it jogs}, N\backslash S: {\it sneezed}\ \Rightarrow\ S$}
The grammaticality proof is:
\disp{\scriptsize
\prooftree
\prooftree
\prooftree
\prooftree
\prooftree
\justifies
N\ \Rightarrow\ N
\endprooftree
\prooftree
\justifies
S\ \Rightarrow\ S
\endprooftree
\justifies
N, N\backslash S\ \Rightarrow\ S
\using \backslash L
\endprooftree
\justifies
[\;], N\backslash S\ \Rightarrow\ S{\uparrow}N
\using {\uparrow}R
\endprooftree
\prooftree
\justifies
\ \Rightarrow\ I
\using IR
\endprooftree
\justifies
N\backslash S\ \Rightarrow\ (S{\uparrow}N){\odot}I
\using {\odot}R
\endprooftree
\prooftree
\prooftree
\justifies
N\ \Rightarrow\ N
\endprooftree
\prooftree
\prooftree
\prooftree
\prooftree
\justifies
N\ \Rightarrow\ N
\endprooftree
\prooftree
\justifies
S\ \Rightarrow\ S
\endprooftree
\justifies
N, N\backslash S\ \Rightarrow\ S
\using \backslash L
\endprooftree
\justifies
[\;], N\backslash S\ \Rightarrow\ S{\uparrow}N
\using {\uparrow}R
\endprooftree
\prooftree
\justifies
S\ \Rightarrow\ S
\endprooftree
\justifies
(S{\uparrow}N){\downarrow}S, N\backslash S\ \Rightarrow\ S
\using {\downarrow}L
\endprooftree
\justifies
N, N\backslash ((S{\uparrow}N){\downarrow}S), N\backslash S\ \Rightarrow\ S
\using \backslash L
\endprooftree
\justifies
N, (N\backslash ((S{\uparrow}N){\downarrow}S))/((S{\uparrow}N){\odot}I), N\backslash S, N\backslash S\ \Rightarrow\ S
\using /L
\endprooftree}
This yields semantics:
\disp{
$[({\it jogs}\ {\it j})\wedge ({\it sneezed}\ {\it j})]$}

\subsection{Parentheticals}

We make the simplifying assumption that a parenthetical adverbial such as
\lingform{fortunately\/} can appear freely. Then our lexical assignment yields the
following series of examples and analyses.
\disp{
${\bf fortunately}{+}{\bf john}{+}{\bf has}{+}{\bf perseverance}: S$}
\disp{
$(S{\uparrow}I){\downarrow}S: \lambda A({\it fortunately}\ ({\it A}\ {\it d})), N: {\it j}, (N\backslash S)/N: {\it has}, N: {\it perseverance}\ \Rightarrow\ S$}
\disp{
\prooftree
\prooftree
\prooftree
\prooftree
\prooftree
\justifies
N\ \Rightarrow\ N
\endprooftree
\prooftree
\prooftree
\justifies
N\ \Rightarrow\ N
\endprooftree
\prooftree
\justifies
S\ \Rightarrow\ S
\endprooftree
\justifies
N, N\backslash S\ \Rightarrow\ S
\using \backslash L
\endprooftree
\justifies
N, (N\backslash S)/N, N\ \Rightarrow\ S
\using /L
\endprooftree
\justifies
I, N, (N\backslash S)/N, N\ \Rightarrow\ S
\using IL
\endprooftree
\justifies
[\;], N, (N\backslash S)/N, N\ \Rightarrow\ S{\uparrow}I
\using {\uparrow}R
\endprooftree
\prooftree
\justifies
S\ \Rightarrow\ S
\endprooftree
\justifies
(S{\uparrow}I){\downarrow}S, N, (N\backslash S)/N, N\ \Rightarrow\ S
\using {\downarrow}L
\endprooftree}
\disp{
$({\it fortunately}\ (({\it has}\ {\it perseverance})\ {\it j}))$}

\disp{
${\bf john}{+}{\bf fortunately}{+}{\bf has}{+}{\bf perseverance}: S$}
\disp{
$N: {\it j}, (S{\uparrow}I){\downarrow}S: \lambda A({\it fortunately}\ ({\it A}\ {\it d})), (N\backslash S)/N: {\it has}, N: {\it perseverance}\ \Rightarrow\ S$}
\disp{
\prooftree
\prooftree
\prooftree
\prooftree
\prooftree
\justifies
N\ \Rightarrow\ N
\endprooftree
\prooftree
\prooftree
\justifies
N\ \Rightarrow\ N
\endprooftree
\prooftree
\justifies
S\ \Rightarrow\ S
\endprooftree
\justifies
N, N\backslash S\ \Rightarrow\ S
\using \backslash L
\endprooftree
\justifies
N, (N\backslash S)/N, N\ \Rightarrow\ S
\using /L
\endprooftree
\justifies
N, I, (N\backslash S)/N, N\ \Rightarrow\ S
\using IL
\endprooftree
\justifies
N, [\;], (N\backslash S)/N, N\ \Rightarrow\ S{\uparrow}I
\using {\uparrow}R
\endprooftree
\prooftree
\justifies
S\ \Rightarrow\ S
\endprooftree
\justifies
N, (S{\uparrow}I){\downarrow}S, (N\backslash S)/N, N\ \Rightarrow\ S
\using {\downarrow}L
\endprooftree}
\disp{
$({\it fortunately}\ (({\it has}\ {\it perseverance})\ {\it j}))$}

\disp{
${\bf john}{+}{\bf has}{+}{\bf fortunately}{+}{\bf perseverance}: S$}
\disp{
$N: {\it j}, (N\backslash S)/N: {\it has}, (S{\uparrow}I){\downarrow}S: \lambda A({\it fortunately}\ ({\it A}\ {\it d})), N: {\it perseverance}\ \Rightarrow\ S$}
\disp{
\prooftree
\prooftree
\prooftree
\prooftree
\prooftree
\justifies
N\ \Rightarrow\ N
\endprooftree
\justifies
I, N\ \Rightarrow\ N
\using IL
\endprooftree
\prooftree
\prooftree
\justifies
N\ \Rightarrow\ N
\endprooftree
\prooftree
\justifies
S\ \Rightarrow\ S
\endprooftree
\justifies
N, N\backslash S\ \Rightarrow\ S
\using \backslash L
\endprooftree
\justifies
N, (N\backslash S)/N, I, N\ \Rightarrow\ S
\using /L
\endprooftree
\justifies
N, (N\backslash S)/N, [\;], N\ \Rightarrow\ S{\uparrow}I
\using {\uparrow}R
\endprooftree
\prooftree
\justifies
S\ \Rightarrow\ S
\endprooftree
\justifies
N, (N\backslash S)/N, (S{\uparrow}I){\downarrow}S, N\ \Rightarrow\ S
\using {\downarrow}L
\endprooftree}
\disp{
$({\it fortunately}\ (({\it has}\ {\it perseverance})\ {\it j}))$}

\disp{${\bf john}{+}{\bf has}{+}{\bf perseverance}{+}{\bf fortunately}: S$}
\disp{
$N: {\it j}, (N\backslash S)/N: {\it has}, N: {\it perseverance}, (S{\uparrow}I){\downarrow}S:\\ \lambda A({\it fortunately}\ ({\it A}\ {\it d}))\ \Rightarrow\ S$}
\disp{
\prooftree
\prooftree
\prooftree
\prooftree
\justifies
N\ \Rightarrow\ N
\endprooftree
\prooftree
\prooftree
\justifies
N\ \Rightarrow\ N
\endprooftree
\prooftree
\prooftree
\justifies
S\ \Rightarrow\ S
\endprooftree
\justifies
S, I\ \Rightarrow\ S
\using IL
\endprooftree
\justifies
N, N\backslash S, I\ \Rightarrow\ S
\using \backslash L
\endprooftree
\justifies
N, (N\backslash S)/N, N, I\ \Rightarrow\ S
\using /L
\endprooftree
\justifies
N, (N\backslash S)/N, N, [\;]\ \Rightarrow\ S{\uparrow}I
\using {\uparrow}R
\endprooftree
\prooftree
\justifies
S\ \Rightarrow\ S
\endprooftree
\justifies
N, (N\backslash S)/N, N, (S{\uparrow}I){\downarrow}S\ \Rightarrow\ S
\using {\downarrow}L
\endprooftree}
\disp{
$({\it fortunately}\ (({\it has}\ {\it perseverance})\ {\it j}))$}

\subsection{Gapping}

\begin{figure}
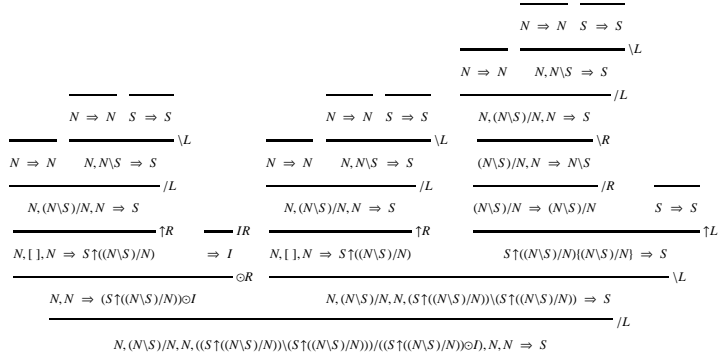

$$
\tiny
\prooftree
\prooftree
\prooftree
\prooftree
\prooftree
\justifies
N\ \Rightarrow\ N
\endprooftree
\prooftree
\prooftree
\justifies
N\ \Rightarrow\ N
\endprooftree
\prooftree
\justifies
S\ \Rightarrow\ S
\endprooftree
\justifies
N, N\backslash S\ \Rightarrow\ S
\using \backslash L
\endprooftree
\justifies
N, (N\backslash S)/N, N\ \Rightarrow\ S
\using /L
\endprooftree
\justifies
N, [\;], N\ \Rightarrow\ S{\uparrow}((N\backslash S)/N)
\using {\uparrow}R
\endprooftree
\prooftree
\justifies
\ \Rightarrow\ I
\using IR
\endprooftree
\justifies
N, N\ \Rightarrow\ (S{\uparrow}((N\backslash S)/N)){\odot}I
\using {\odot}R
\endprooftree
\prooftree
\prooftree
\prooftree
\prooftree
\justifies
N\ \Rightarrow\ N
\endprooftree
\prooftree
\prooftree
\justifies
N\ \Rightarrow\ N
\endprooftree
\prooftree
\justifies
S\ \Rightarrow\ S
\endprooftree
\justifies
N, N\backslash S\ \Rightarrow\ S
\using \backslash L
\endprooftree
\justifies
N, (N\backslash S)/N, N\ \Rightarrow\ S
\using /L
\endprooftree
\justifies
N, [\;], N\ \Rightarrow\ S{\uparrow}((N\backslash S)/N)
\using {\uparrow}R
\endprooftree
\prooftree
\prooftree
\prooftree
\prooftree
\prooftree
\justifies
N\ \Rightarrow\ N
\endprooftree
\prooftree
\prooftree
\justifies
N\ \Rightarrow\ N
\endprooftree
\prooftree
\justifies
S\ \Rightarrow\ S
\endprooftree
\justifies
N, N\backslash S\ \Rightarrow\ S
\using \backslash L
\endprooftree
\justifies
N, (N\backslash S)/N, N\ \Rightarrow\ S
\using /L
\endprooftree
\justifies
(N\backslash S)/N, N\ \Rightarrow\ N\backslash S
\using \backslash R
\endprooftree
\justifies
(N\backslash S)/N\ \Rightarrow\ (N\backslash S)/N
\using /R
\endprooftree
\prooftree
\justifies
S\ \Rightarrow\ S
\endprooftree
\justifies
S{\uparrow}((N\backslash S)/N)\{(N\backslash S)/N\}\ \Rightarrow\ S
\using {\uparrow}L
\endprooftree
\justifies
N, (N\backslash S)/N, N, (S{\uparrow}((N\backslash S)/N))\backslash (S{\uparrow}((N\backslash S)/N))\ \Rightarrow\ S
\using \backslash L
\endprooftree
\justifies
N, (N\backslash S)/N, N, ((S{\uparrow}((N\backslash S)/N))\backslash (S{\uparrow}((N\backslash S)/N)))/((S{\uparrow}((N\backslash S)/N)){\odot}I), N, N\ \Rightarrow\ S
\using /L
\endprooftree
$$
\caption{\lingform{John studies logic, and Charles, phonetics}}
\label{gapping}
\end{figure}

In gapping coordination a verb in the left conjunct is understood in the
right conjunct:
\disp{
${\bf john}{+}{\bf studies}{+}{\bf logic}{+}{\bf and}{+}{\bf charles}{+}{\bf phonetics}: S$}
Lexical lookup for the gapping coordinator type yields:
\disp{
$N: {\it j}, (N\backslash S)/N: {\it studies}, N: {\it logic}, ((S{\uparrow}((N\backslash S)/N))\backslash (S{\uparrow}\\((N\backslash S)/N)))/((S{\uparrow}((N\backslash S)/N)){\odot}I): \lambda A\lambda B\lambda C[({\it B}\ {\it C})\wedge\\ (\pi_1{\it A}\ {\it C})], N: {\it c}, N: {\it phonetics}\ \Rightarrow\ S$}
The derivation is as shown in Figure~\ref{gapping}.
This yields semantics:
\disp{
$[(({\it studies}\ {\it logic})\ {\it j})\wedge (({\it studies}\ {\it phonetics})\ {\it c})]$}

\subsection{Comparative Subdeletion} 

In comparative subdeletion a clause containing a comparative determiner
such as \lingform{more} is compared to a \lingform{than}-clause from which
a determiner is missing, with the comparative semantics:
\disp{
${\bf john}{+}{\bf ate}{+}{\bf more}{+}{\bf donuts}{+}{\bf than}{+}{\bf mary}{+}{\bf bought}{+}{\bf bagels}: S$}
For this example lexical lookup of our assignments yields:
\disp{
$N: {\it j}, (N\backslash S)/N: {\it ate}, (S{\uparrow}(((S{\uparrow}N){\downarrow}S)/{\it CN})){\downarrow}(S/(({\it CP}{\uparrow}\\(((S{\uparrow}N){\downarrow}S)/{\it CN})){\odot}I)): \lambda A\lambda B[|\lambda C({\it A}\ \lambda D\lambda E[({\it D}\ {\it C})\wedge\\ ({\it E}\ {\it C})])|>|\lambda C(\pi_1{\it B}\ \lambda D\lambda E[({\it D}\ {\it C})\wedge ({\it E}\ {\it C})])|], {\it CN}: {\it donuts},\\ {\it CP}/S: \lambda A{\it A}, N: {\it m}, (N\backslash S)/N: {\it bought}, {\it CN}: {\it bagels}\ \Rightarrow\ S$}
A sequent proof derivation is given in Figure~\ref{bagels}.
\begin{figure}
$$
\rotatebox{-90}{\scriptsize
\prooftree
\prooftree
\prooftree
\prooftree
\justifies
{\it CN}\ \Rightarrow\ {\it CN}
\endprooftree
\prooftree
\prooftree
\prooftree
\prooftree
\justifies
N\ \Rightarrow\ N
\endprooftree
\prooftree
\prooftree
\justifies
N\ \Rightarrow\ N
\endprooftree
\prooftree
\justifies
S\ \Rightarrow\ S
\endprooftree
\justifies
N, N\backslash S\ \Rightarrow\ S
\using \backslash L
\endprooftree
\justifies
N, (N\backslash S)/N, N\ \Rightarrow\ S
\using /L
\endprooftree
\justifies
N, (N\backslash S)/N, [\;]\ \Rightarrow\ S{\uparrow}N
\using {\uparrow}R
\endprooftree
\prooftree
\justifies
S\ \Rightarrow\ S
\endprooftree
\justifies
N, (N\backslash S)/N, (S{\uparrow}N){\downarrow}S\ \Rightarrow\ S
\using {\downarrow}L
\endprooftree
\justifies
N, (N\backslash S)/N, ((S{\uparrow}N){\downarrow}S)/{\it CN}, {\it CN}\ \Rightarrow\ S
\using /L
\endprooftree
\justifies
N, (N\backslash S)/N, [\;], {\it CN}\ \Rightarrow\ S{\uparrow}(((S{\uparrow}N){\downarrow}S)/{\it CN})
\using {\uparrow}R
\endprooftree
\prooftree
\prooftree
\prooftree
\prooftree
\prooftree
\prooftree
\justifies
{\it CN}\ \Rightarrow\ {\it CN}
\endprooftree
\prooftree
\prooftree
\prooftree
\prooftree
\justifies
N\ \Rightarrow\ N
\endprooftree
\prooftree
\prooftree
\justifies
N\ \Rightarrow\ N
\endprooftree
\prooftree
\justifies
S\ \Rightarrow\ S
\endprooftree
\justifies
N, N\backslash S\ \Rightarrow\ S
\using \backslash L
\endprooftree
\justifies
N, (N\backslash S)/N, N\ \Rightarrow\ S
\using /L
\endprooftree
\justifies
N, (N\backslash S)/N, [\;]\ \Rightarrow\ S{\uparrow}N
\using {\uparrow}R
\endprooftree
\prooftree
\justifies
S\ \Rightarrow\ S
\endprooftree
\justifies
N, (N\backslash S)/N, (S{\uparrow}N){\downarrow}S\ \Rightarrow\ S
\using {\downarrow}L
\endprooftree
\justifies
N, (N\backslash S)/N, ((S{\uparrow}N){\downarrow}S)/{\it CN}, {\it CN}\ \Rightarrow\ S
\using /L
\endprooftree
\prooftree
\justifies
{\it CP}\ \Rightarrow\ {\it CP}
\endprooftree
\justifies
{\it CP}/S, N, (N\backslash S)/N, ((S{\uparrow}N){\downarrow}S)/{\it CN}, {\it CN}\ \Rightarrow\ {\it CP}
\using /L
\endprooftree
\justifies
{\it CP}/S, N, (N\backslash S)/N, [\;], {\it CN}\ \Rightarrow\ {\it CP}{\uparrow}(((S{\uparrow}N){\downarrow}S)/{\it CN})
\using {\uparrow}R
\endprooftree
\prooftree
\justifies
\ \Rightarrow\ I
\using IR
\endprooftree
\justifies
{\it CP}/S, N, (N\backslash S)/N, {\it CN}\ \Rightarrow\ ({\it CP}{\uparrow}(((S{\uparrow}N){\downarrow}S)/{\it CN})){\odot}I
\using {\odot}R
\endprooftree
\prooftree
\justifies
S\ \Rightarrow\ S
\endprooftree
\justifies
S/(({\it CP}{\uparrow}(((S{\uparrow}N){\downarrow}S)/{\it CN})){\odot}I), {\it CP}/S, N, (N\backslash S)/N, {\it CN}\ \Rightarrow\ S
\using /L
\endprooftree
\justifies
N, (N\backslash S)/N, (S{\uparrow}(((S{\uparrow}N){\downarrow}S)/{\it CN})){\downarrow}(S/(({\it CP}{\uparrow}(((S{\uparrow}N){\downarrow}S)/{\it CN})){\odot}I)), {\it CN}, {\it CP}/S, N, (N\backslash S)/N, {\it CN}\ \Rightarrow\ S
\using {\downarrow}L
\endprooftree}
$$
\caption{John ate more donuts than Mary bought bagels}
\label{bagels}
\end{figure}
This yields semantics:
\disp{
$[|\lambda C[({\it donuts}\ {\it C})\wedge (({\it ate}\ {\it C})\ {\it j})]|>|\lambda C[({\it bagels}\ {\it C})\wedge\\ (({\it bought}\ {\it C})\ {\it m})]|]$}

\subsection{Reflexivization}

In our example the reflexive receives its interpretation from the subject:
\disp{
${\bf john}{+}{\bf sent}{+}{\bf himself}{+}{\bf flowers}: S$}
Lexical lookup yields:
\disp{
$N: {\it j}, (N\backslash S)/(N{\bullet}N): \lambda A(({\it sent}\ \pi_1{\it A})\ \pi_2{\it A}),\\ ((N\backslash S){\uparrow}N){\downarrow}(N\backslash S): \lambda A\lambda B(({\it A}\ {\it B})\ {\it B}), N: {\it flowers}\ \Rightarrow\ S$}
This has derivation:
\disp{\footnotesize
\prooftree
\prooftree
\prooftree
\prooftree
\prooftree
\prooftree
\justifies
N\ \Rightarrow\ N
\endprooftree
\prooftree
\justifies
N\ \Rightarrow\ N
\endprooftree
\justifies
N, N\ \Rightarrow\ N{\bullet}N
\using \bullet R
\endprooftree
\prooftree
\prooftree
\justifies
N\ \Rightarrow\ N
\endprooftree
\prooftree
\justifies
S\ \Rightarrow\ S
\endprooftree
\justifies
N, N\backslash S\ \Rightarrow\ S
\using \backslash L
\endprooftree
\justifies
N, (N\backslash S)/(N{\bullet}N), N, N\ \Rightarrow\ S
\using /L
\endprooftree
\justifies
(N\backslash S)/(N{\bullet}N), N, N\ \Rightarrow\ N\backslash S
\using \backslash R
\endprooftree
\justifies
(N\backslash S)/(N{\bullet}N), [\;], N\ \Rightarrow\ (N\backslash S){\uparrow}N
\using {\uparrow}R
\endprooftree
\prooftree
\prooftree
\justifies
N\ \Rightarrow\ N
\endprooftree
\prooftree
\justifies
S\ \Rightarrow\ S
\endprooftree
\justifies
N, N\backslash S\ \Rightarrow\ S
\using \backslash L
\endprooftree
\justifies
N, (N\backslash S)/(N{\bullet}N), ((N\backslash S){\uparrow}N){\downarrow}(N\backslash S), N\ \Rightarrow\ S
\using {\downarrow}L
\endprooftree}
This delivers semantics:
\disp{
$((({\it sent}\ {\it j})\ {\it flowers})\ {\it j})$}

\section{Cut-Elimination}

\label{cutsect}

\cite{lambek:mathematics} proved Cut-elimination for the Lambek calculus \AL.
Cut-elimination states that every theorem can be proved without the use of Cut.
Lambek's proof is simpler than that of Gentzen for standard logic
due to the absence of structural rules.
It consists of defining a notion of degree of Cut instances and showing how
Cuts in a proof can be succesively replaced by Cuts of lower degree
until they are removed altogether.
Thus Lambek's proof provides an algorithm for transforming proofs
into Cut-free counterparts.
The Cut-elimination theorem has as corollaries the subformula property
and decidability.

Here we prove Cut-elimination for the displacement calculus {\bf D}.
Like \AL{}, {\bf D} contains no structural rules (structural properties are built
into the sequent calculus notation) and the Cut-elimination is proved
following the same strategy as for {\bf L}.

We define the \techterm{weight\/} $|A|$ of a type $A$ as the number
of connectives occurrences (including units) that it contains.
The weight $|\Gamma|$ of a configuration is the sum of the weights
of the types that occur in it, that is, it is defined recursively as follows:
\disp{$
\begin{array}[t]{lll}
|\Lambda| & = & 0\\
|\sep| & = & 0\\
|A| & = & |A| \\
|A\{\Gamma_1:\cdots:\Gamma_{i+1}\}| & = & |A|+\displaystyle\sum_{j=1}^{i+1}|\Gamma_j|\\
|\Gamma, \Theta| & = & |\Gamma|+|\Theta|
\end{array}
$}
The weight of a hypercontext is defined similarly with a hole having weight zero.

Consider the Cut rule:
\disp{$
\prooftree \Gamma\yields A \tb \Delta\langle \vect A\rangle\yields B\justifies \Delta\langle \Gamma\rangle\yields B \using Cut \ \ (\star)\endprooftree
$\label{cut}}
We define the \techterm{degree} $d(\star)$ of an instance $\star$ of the Cut rule as follows:
\disp{
$d(\star) = |\Gamma|+|\Delta|+ |A|+|B|$}
We call the type $A$ in (\ref{cut}) the \techterm{Cut formula}.
We call the type which is newly created by a logical rule the \techterm{active formula}.
Consider a proof which is not Cut-free. 
Then there is some Cut-instance above which there are no Cuts.
We will show that this Cut can either be removed or replaced by
one or two Cuts of lower degree. The following three cases are exhaustive:
\disp{
\begin{itemize}
\item
A premise of the Cut is the identity axiom: then the conclusion is identical to the
other premise and the Cut as a whole can be removed.
\item
Both the premises are conclusions of logical rules and it is not the case
that the Cut formula is the active 
formula of both premises: then we apply \techterm{permutation conversion} cases.
\item
Both the premises are conclusions of logical rules and the Cut formula is
the active formula of both premises: then we apply \techterm{principal Cut\/} cases.
\end{itemize}}
There are several cases to consider. We give representative examples.

\subsection{Permutation conversion cases}

\subsubsection{The active formula in the left premise of the Cut rule is not the Cut formula}

\begin{itemize}

\item The rule applying at the left premise of the Cut rule is $\dprod_i\, L$:
\vtab
\prooftree \prooftree \Delta\langle\vect B|_i\vect C\rangle\yields A \justifies \Delta\langle\vect{B\dprod_i C}\rangle\yields A
\using \dprod_i L\endprooftree \tb \Gamma\langle\vect A\rangle\yields D \justifies \Gamma\langle\Delta\langle\vect{B\dprod_i C}\rangle
\rangle\yields 
D\using Cut\endprooftree 
$$\leadsto$$
\prooftree \prooftree \Delta\langle\vect B|_i\vect C\rangle\yields  A  \tb \Gamma\langle \vect A\rangle \yields D \justifies \Gamma\langle\Delta\langle\vect B|_i\vect C\rangle\rangle
\yields  A \using Cut\endprooftree \justifies \Gamma\langle\Delta\langle\vect{B\dprod_i C}\rangle\rangle\yields D\using 
\dprod_i L\endprooftree\vspace{0.5cm}

\item The rule applying at the left premise of the Cut rule is $\extract_i{} L$:
\vtab
\prooftree \prooftree \Gamma\langle\vect C\rangle\yields  A \tb \Delta\yields B \justifies \Gamma\langle\vect{C\extract_i B}|_i\Delta\rangle
\yields A\using \extract_i L\endprooftree \tb \Theta\langle\vect A\rangle\yields  D\justifies \Theta\langle\Gamma\langle\vect{C\extract_i B}|_i\Delta\rangle\rangle
\yields  D\using Cut\endprooftree
$$\leadsto$$
\prooftree \prooftree \Gamma\langle\vect C\rangle\yields  A \tb \Theta\langle\vect A\rangle\yields D \justifies 
\Theta\langle\Gamma\langle\vect C\rangle\rangle\yields  D\using Cut\endprooftree \tb \Delta\yields B \justifies \Theta\langle\Gamma\langle\vect{C\extract_i B}|_i\Delta\rangle\rangle
\yields D\using \extract_i L\endprooftree

\item The rule applying at the left premise of the Cut rule is $JL$:
\vtab
\prooftree \prooftree \Gamma\langle\sep\rangle\yields A \justifies\Gamma\langle\vect J\rangle\yields A \using JL\endprooftree \tb \Delta\langle\vect A\rangle\yields B\justifies\Delta\langle\vect A\rangle\yields B \justifies\Delta\langle\Gamma\langle\vect J\rangle\rangle\yields B\using Cut\endprooftree
$$\leadsto$$
\prooftree  \prooftree \Gamma\langle\sep\rangle\yields A \tb \Delta\langle\vect A\rangle\yields B \justifies\Delta\langle\Gamma\langle\sep\rangle\rangle\yields A\using Cut\endprooftree \justifies\Delta\langle\Gamma\langle\vect J\rangle\rangle\yields B \using JL\endprooftree

\end{itemize}

\subsubsection{The active formula in the right premise of the Cut rule is not the Cut formula}

\begin{itemize}
\item The rule applying at the right premise of the Cut rule is $\extract_i\, L$ :
\vtab
\prooftree \Delta\yields  A \tb \prooftree \Gamma\langle\vect A;\vect C\rangle\yields D\tb 
\Theta\yields B \justifies \Gamma\langle\vect A;\vect{C\extract_i B}|_i\Theta\rangle\yields D\using \extract_i L\endprooftree \justifies \Gamma\langle\Delta;\vect{C\extract_i B}|_i\Theta\rangle
\yields  D\using Cut\endprooftree
$$\leadsto$$
\prooftree \prooftree \Delta\yields A\tb \Gamma\langle\vect A;\vect C\rangle\yields D \justifies 
\Gamma\langle\Delta;\vect C\rangle\yields D\using Cut\endprooftree \tb \Theta\yields B \justifies \Gamma\langle\Delta;\vect{C\extract_i B}|_i\Theta\rangle
\yields D\using \extract_i L\endprooftree

\item The rule applying at the right premise of the Cut rule is $\extract_i\,R$:
\vtab
\prooftree \Delta\yields A \tb \prooftree \Gamma\langle\vect A\rangle|_i\vect B\yields C \justifies \Gamma\langle\vect A\rangle\yields
 C\extract_i B\using \extract_i R\endprooftree \justifies \Gamma\langle\Delta\rangle\yields C\extract_i B\using Cut
\endprooftree
$$\leadsto$$
\prooftree\prooftree  \Delta\yields A \tb \Gamma\langle\vect A\rangle|_i\vect B\yields  C \justifies 
\Gamma\langle\Delta\rangle|_i\vect B\yields  C\using Cut\endprooftree\justifies \Gamma\langle\Delta\rangle\yields
C\extract_i B\using \extract_i R\endprooftree

\item The rule applying at the right premise of the Cut rule is $\dprod_i\,L$:
\vtab
\prooftree \Delta\yields A \tb \prooftree \Gamma\langle\vect A;\vect B|_i \vect C\rangle\yields D \justifies \Gamma\langle\vect A;\vect{B\dprod_i C}\rangle 
\yields D\using \dprod_i L\endprooftree \justifies \Gamma\langle\Delta;\vect{B\dprod_i C}\rangle \yields D\using Cut\endprooftree
$$\leadsto$$
\prooftree \prooftree \Delta\yields A \tb  \Gamma\langle\vect A;\vect B|_i\vect C\rangle \yields D\justifies \Gamma\langle\Delta;\vect B|_i\vect C\rangle 
\yields D\using Cut\endprooftree \justifies \Gamma\langle\Delta;\vect{B\dprod_i C}\rangle 
\yields D\using \dprod_i L\endprooftree

\item The rule applying at the right premise of the Cut rule is $\dprod_i\,R$:
\vtab
\prooftree \Delta\yields A \tb \prooftree \Gamma\langle\vect A\rangle\yields B\tb \Theta\yields C 
\justifies \Gamma\langle\vect A\rangle|_i\Theta\yields B\dprod_i C\using \dprod_i R\endprooftree \justifies \Gamma\langle\Delta\rangle|_i\Theta
\yields B\dprod_i C\using Cut\endprooftree
$$\leadsto$$
\prooftree \prooftree \Delta\yields A \tb \Gamma\langle\vect A\rangle\yields B\justifies 
\Gamma\langle\Delta\rangle\yields B\using Cut\endprooftree \tb \Theta\yields C \justifies \Gamma\langle\Delta\rangle|_i\Theta
\yields B\dprod_i C\using \dprod_i R\endprooftree

\end{itemize}

\subsection{Principal Cut cases}

\begin{itemize}
\item The rules applying at the left and right premises of the Cut rule are respectively $\dprod_i\,R$ and $\dprod_i\,L$:
\vtab
\prooftree  \prooftree \Delta\yields A \tb \Gamma\yields B \justifies \Delta|_i\Gamma\yields A\dprod_i B \using \dprod_i R\endprooftree\tb \prooftree \Theta\langle\vect A|_i\vect B\rangle
\yields C \justifies \Theta\langle\vect{A\dprod_i B}\rangle\yields C\using \dprod_i L\endprooftree\justifies \Theta\langle\Delta|_i\Gamma\rangle
\yields C\using Cut\endprooftree
$$\leadsto$$
\prooftree \Gamma\yields B \prooftree \Delta\yields A \tb \Theta\langle\vect A|_i\vect B\rangle\yields C\justifies \Theta\langle\Delta|_i\vect B\rangle
\yields C\using Cut\endprooftree\justifies \Theta\langle\Delta|_i\Gamma\rangle\yields C\using Cut\endprooftree

\item The rules applying at the left and right premises of the Cut rule are respectively $\extract_i\,R$ and $\extract_i\,L$:
\vtab
\prooftree  \prooftree \Delta|_i\vect A\yields  B \justifies \Delta\yields B\extract_i A\using \extract_i R\endprooftree \tb  \prooftree \Gamma
\yields A \tb \Theta\langle\vect B\rangle\yields  C \justifies \Theta\langle B\extract_i A|_i\Gamma\rangle\yields 
C\using \extract_i L\endprooftree \justifies \Theta\langle\Delta|_i\Gamma\rangle\yields C\using Cut\endprooftree
$$\leadsto$$
\prooftree \Delta\yields A \tb \prooftree \Delta|_i\vect A\yields  B \tb \Theta\langle\vect B\rangle
\yields C\justifies \Theta\langle\vect A|_i\Gamma\rangle\yields C\using Cut\endprooftree \justifies \Theta\langle\Delta|_i\Gamma\rangle
\yields C\using Cut\endprooftree

\item The rules applying at the left and right premises of the Cut rule are respectively
$IR$ and $IL$:
\vtab
\prooftree \prooftree  \justifies\Lambda\yields I \using IR\endprooftree \tb \prooftree \Delta\langle\Lambda\rangle\yields A \justifies\Delta\langle I\rangle\yields A \using IL\endprooftree \justifies \Delta\langle\Lambda\rangle\yields A\using Cut\endprooftree
$$\leadsto$$
$\Delta\langle\Lambda\rangle\yields A$

\item The rules applying at the left and right premises of the Cut rule are respectively
$JR$ and $JL$:
\vtab
\prooftree \prooftree  \justifies\sep\yields J \using JR\endprooftree \tb \prooftree \Delta\langle\sep\rangle\yields A \justifies\Delta\langle \vect J\rangle\yields A \using JL\endprooftree \justifies \Delta\langle\sep\rangle\yields A\using Cut\endprooftree
$$\leadsto$$
$\Delta\langle\sep\rangle\yields A$

\end{itemize}

\section{Conclusion}

\label{conclsect}

The reasoning given in the previous section yields the following properties:

\mytheorem{Cut-elimination for {\bf D}}{Every theorem 
of the displacement calculus {\bf D} has a Cut-free proof.}
\prf{As we have indicated, in every proof which is not Cut-free
it is always possible to replace a Cut
above which there are no Cuts either by replacing it by one or two Cuts
of lower degree or by removing it altogether, conserving the endsequent
of the proof.
Since the degree of a Cut is always finite and non-negative,
repeated application of this procedure will transform every proof
into a Cut-free counterpart.}

\mycorollary{Subformula property for {\bf D}}{Every theorem of the
displacement calculus {\bf D} has a proof in which appear only
subformulas of the theorem.}
\prf{In every rule except Cut every formula in a premise is a subformula
of a formula in the conclusion,
and Cut itself is eliminable. 
Hence, every theorem has a proof containing only subformulas of the
theorem, namely any one of its Cut-free proofs.}

\mycorollary{Decidability of {\bf D}}{It is decidable whether a (hyper)sequent of
{\bf D} is a theorem.}
\prf{In backward chaining Cut-free hypersequent proof search a hypersequent
can be matched against a rule only in a finite number of ways and generates
only a finite number of subgoals. Hence the backward chaining 
Cut-free hypersequent proof search space
is finite and it is determined in finite time whether a sequent is a theorem.}

This paper offers an account of generalized discontinuity in the sense anticipated
in \cite{morrill:merenciano} in respect of sorts and in
\cite{morrill:02} in respect of unboundedly many positions of discontinuity.
All the applications of Section~\ref{outsect} fall within the fragment
with just one point of discontinuity but the full calculus allows
arbitrarily many such points.\footnote{The sequent notation here
employs an improvement over that of \cite{mfv:iwcs07} following a suggestion by Sylvain Salvati (p.c.).}
The program of generalizing categorial grammar in this way goes
back to \cite{moortgat:phd} and \cite{bach81}.

Logically,
we have generalized and extended the concatenative multiplicative
connectives of Lambek calculus/intuitionistic non-commutative linear logic
with families of non-concatenative multiplicative connectives,
but concatenation remains the unique primitive mode of composition
and the calculus remains free of structural rules.
These features contribute to the simplicity of implementation
of displacement calculus parsing-as-deduction.

{
\bibliography{bib100414}
}

\end{document}